# Instruct-ICL: Instruction-Guided In-Context Learning for Post-Disaster Damage Assessment

Armin Zarbaft
*Lehigh Uniersity*
Pennsylvania, USA
arz328@lehigh.edu

Ehsan Karimi
*Lehigh Uniersity*
Pennsylvania, USA
ehk224@lehigh.edu

Nhut Le
*Lehigh Uniersity*
Pennsylvania, USA
nhl224@lehigh.edu

Maryam Rahnemoonfar*
*Lehigh Uniersity*
Pennsylvania, USA
maryam@lehigh.edu
*Corresponding author

***Abstract*—Rapid and accurate situational awareness is essential for effective response during natural disasters, where delays in analysis can significantly hinder decision-making. Training task-specific models for post-disaster assessment is often time-consuming and computationally expensive, making such approaches impractical in time-critical scenarios. Consequently, pretrained multimodal large language models (MLLMs) have emerged as a promising alternative for post-disaster visual question answering (VQA), a task that aims to answer structured questions about visual scenes by jointly reasoning over images and text. While these models demonstrate strong multimodal reasoning capabilities, their responses can be sensitive to prompt formulation, which can limit their reliability in real-world disaster assessment scenarios. In this paper, we investigate whether structured reasoning strategies can improve the reliability of pretrained MLLMs for post-disaster VQA. Specifically, we explore multiple prompting paradigms in which one MLLM is used to generate task-specific instructions that serve as Chain-of-Thought (CoT) guidance for a second MLLM. These instructions are incorporated during answer generation with varying degrees of in-context learning (ICL), enabling the model to leverage both explicit reasoning guidance and contextual examples. We conduct our evaluation on the FloodNet dataset and compare these approaches against a zero-shot baseline. Our results demonstrate that integrating instruction-driven CoT reasoning consistently improves answer accuracy.**

## I. Introduction

RAPID and informed response following a natural disaster is critical for minimizing loss of life and long-term damage. Numerous studies emphasize that the effectiveness of rescue and recovery operations depends heavily on how quickly responders can assess the situation on the ground and allocate resources accordingly [1] [2]. Delays in situational awareness can significantly hinder rescue efforts in large-scale disasters. As a result, there is a growing demand for technologies that can provide timely, accurate, and actionable information immediately after a disaster occurs.

Unmanned Aerial Vehicles (UAVs) have emerged as a powerful tool to address this need by enabling rapid data acquisition over affected areas without placing responders at risk. UAVs can capture detailed aerial imagery shortly after a disaster, offering a comprehensive view of damage patterns, flooded regions, and inaccessible areas. Compared to satellite imagery, UAVs provide higher spatial resolution and greater deployment flexibility, making them particularly suitable for time-critical disaster response scenarios. Recent advances in UAV-based remote sensing have further improved data quality and coverage, allowing emergency managers to obtain near real-time imagery that supports faster and more informed decision-making [3], [4].

To transform aerial imagery into actionable information, frameworks capable of jointly reasoning over visual and textual data are required. Visual Question Answering (VQA) has emerged as a promising paradigm for extracting structured information from images by enabling systems to answer natural-language questions about visual content [5]. Unlike traditional computer vision approaches that focus on fixed classification, VQA provides a flexible interface capable of addressing a wide range of semantic queries from a single image. This adaptability makes VQA particularly appealing for disaster response. Prior studies have demonstrated that VQA can effectively answer diverse questions about post-disaster scenes that are not easily addressed by conventional methods such as convolutional neural networks or cascade classifiers [6]. By jointly reasoning over visual and textual inputs, VQA systems offer a practical way to extract human-interpretable information from post-disaster imagery, highlighting their potential utility in real-world disaster scenarios.

Despite these advantages, many VQA systems rely on fully supervised learning that requires large labeled datasets and extensive training time [7]. In the context of natural disasters, this reliance poses a significant limitation, as labeled datasets may not be readily available for new or rare disaster events, and training or fine-tuning models can be prohibitively time-consuming. Another key challenge arises from the limited generalizability of models trained on narrowly scoped datasets. Models trained primarily on flood or hurricane imagery, for example, may struggle to accurately interpret damage caused by other disaster types due to differences in visual damage patterns and underlying physical mechanisms [8].

To address these limitations, recent research has shifted toward zero-shot methods using pretrained multimodal large language models (MLLMs). These approaches eliminate the need for task-specific training by exploiting the broad knowledge and reasoning capabilities encoded in pretrained models [9]. However, despite strong reasoning abilities, zero-shot MLLMs often suffer from hallucination, producing incorrect answers despite seemingly valid reasoning, which is particularly problematic in high-stakes applications like disaster response. Recent studies have shown that structuring model reasoning through instruction generation and explicitly using these instructions as chain-of-thought (CoT) guidance for a second MLLM can significantly reduce hallucinations and improve answer reliability [10].

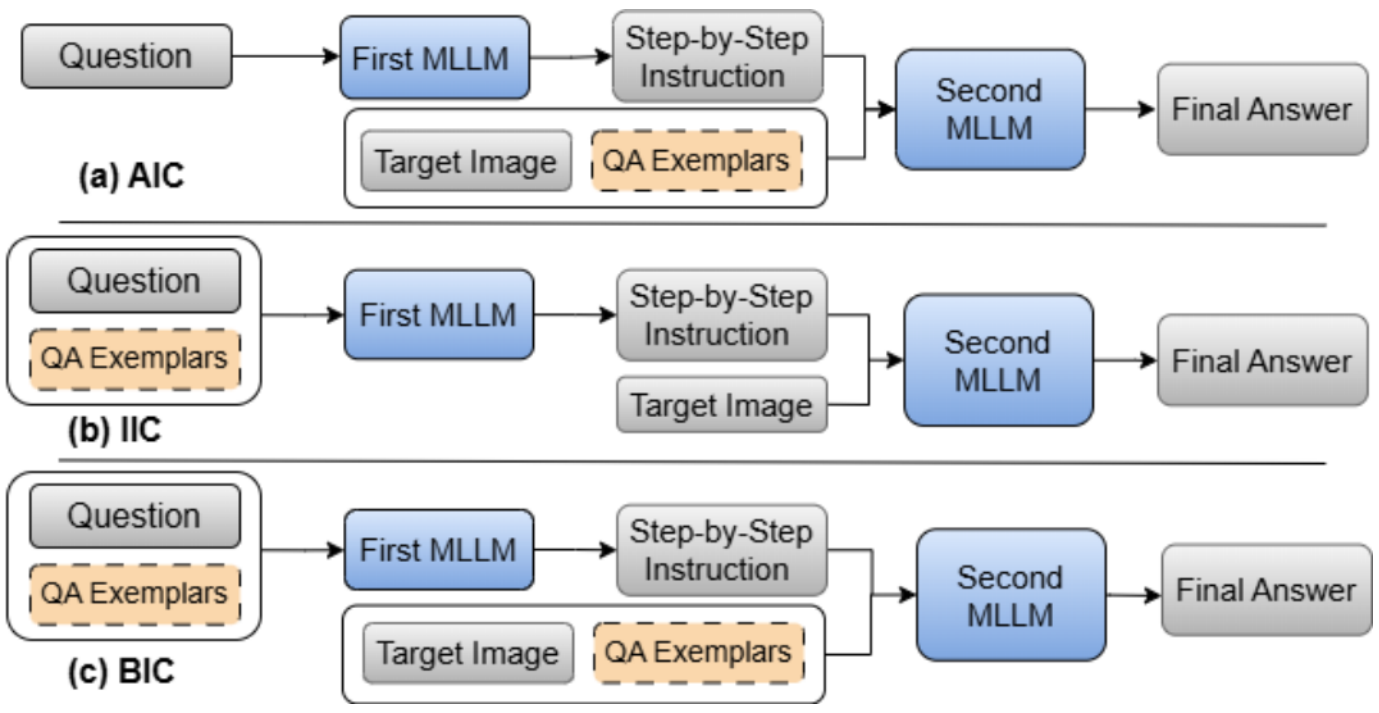

Fig. 1: Illustration of three ICL integration methods withing a two-stage CoT MLLM pipeline

Motivated by these findings, we explore different ways in which ICL can be integrated within a CoT framework to improve the performance of VQA models on the FloodNet dataset [11]. All proposed approaches follow a two-stage MLLM pipeline. In the first stage, an instruction is generated either by providing only the question and prompting the MLLM to produce a step-by-step reasoning instruction, or by additionally supplying similar QA exemplars for in-context learning. In the second stage, the generated instruction is combined with the UAV image to produce the final answer. As illustrated in Fig.1, we evaluate three distinct methods that differ in how ICL is applied across the two stages:

- Answer-ICL CoT (AIC): The first MLLM receives only the question and generates a step-by-step instruction. This instruction is then passed to a second MLLM together with the target image and a set of similar exemplars used for in-context learning to generate the final answer.
- Instruction-ICL CoT (IIC): In-context learning is applied during instruction generation by providing similar exemplars alongside the question to the first MLLM. The resulting instruction is then supplied to a second MLLM together with the target image to produce the final answer.
- Both-ICL CoT (BIC): The same set of similar exemplars is provided to both the instruction-generation MLLM and the answer-generation MLLM, enabling in-context learning at both stages of the pipeline.

The remainder of this paper is organized as follows. Section II reviews related work on post-disaster assessment, VQA, and MLLMs. Section III describes the FloodNet dataset. Section IV presents the proposed methodology in detail, including the different instruction-generation and in-context learning strategies evaluated. Finally, Section V concludes the paper and discusses future research directions.

## II. Related Works

This section briefly reviews prior work on post-disaster assessment, covering semantic segmentation methods, VQA frameworks, and challenges in applying CoT and ICL in MLLMs.

### A. From Semantic Segmentation to VQA-Based Post-Disaster Damage Assessment

Early learning-based approaches to post-disaster damage assessment have predominantly relied on supervised models, with semantic segmentation emerging as one of the most widely adopted techniques for analyzing aerial and UAV imagery. By assigning a semantic label to each pixel, these methods enable fine-grained localization of disaster-related features such as flooded regions and damaged structures [12]–[16]. Such models have been successfully applied to generate detailed spatial damage maps and support situational awareness during emergency response. However, their effectiveness depends on large labeled datasets and task-specific training, which can be costly and time-consuming. In addition, their rigid output structures limit flexibility, as they are typically designed for a narrow set of predefined classes.

To address these limitations, Visual Question Answering (VQA) has emerged as a more flexible, question-driven paradigm for post-disaster assessment [7], [17]. Rather than producing fixed outputs, VQA models can answer diverse, human-interpretable questions about an image without requiring task-specific output heads [18]. Recent work such as ThiFAN-VQA demonstrates the potential of reasoning-based VQA frameworks by bridging the gap between fully supervised and zero-shot approaches. ThiFAN-VQA introduces a two-stage reasoning process in which a generative model produces a structured Chain-of-Thought reasoning trace and directly assigns the final answer within a single prompt, jointly leveraging CoT prompting and in-context learning [19]. Experiments on datasets such as FloodNet show improved accuracy and interpretability compared to baseline VQA methods. These findings highlight the promise of reasoning-oriented VQA for disaster assessment, while also revealing ongoing challenges related to prompt design and reasoning reliability.

### B. Challenges of CoT and ICL in MLLMs

CoT has been shown to significantly enhance the reasoning capabilities of MLLMs, leading to improved accuracy and interpretability across a wide range of tasks [20]. By encouraging models to generate intermediate reasoning steps prior to producing a final answer, CoT enables more structured and transparent decision-making and has been widely validated as an effective prompting strategy for complex reasoning problems. However, recent studies have revealed that combining CoT and ICL within the same prompt can reduce overall performance [21]. Specifically, when MLLMs are simultaneously encouraged to follow exemplar-based patterns while generating explicit reasoning, conflicting cognitive demands may arise, leading to degraded accuracy. Recent work addresses this issue by separating textual reasoning from visual input, demonstrating that “thinking before looking” enables more structured cognitive steps [10]. By first analyzing the question independently and constructing a reasoning plan, the model is less likely to hallucinate visual details that align with learned patterns rather than actual image content. These findings motivate approaches that separate reasoning from image-conditioned inference. Building on these insights, our work explores multiple ways of integrating CoT and ICL across different stages of a VQA pipeline, to improve accuracy and robustness.

## III. Data

For this study, we utilize FloodNet [11], a high-resolution dataset specifically designed for post-disaster damage assessment using UAV imagery. The dataset was captured following Hurricane Harvey in 2017 across Fort Bend County, Texas. It comprises 2,188 high-resolution images paired with 7,355 QA pairs, providing a robust framework for multi-modal scene understanding in disaster scenarios. The VQA component of FloodNet is categorized into four distinct types of reasoning tasks, designed to evaluate a model's ability to interpret complex environmental damage:

- Counting: This category is subdivided into two levels of difficulty:
  - Simple Counting: Requires the model to identify the total number of buildings within the frame.
  - Complex Counting: Requires conditional reasoning to count buildings that satisfy specific criteria, such as "flooded" or "non-flooded."
- Condition Recognition: These tasks assess the flooding status of various infrastructure elements. Questions target the overall scene, roads, or buildings. They are formatted either as binary yes/no queries (e.g., "Is the road easily accessible?") or as binary flooded/non-flooded classifications (e.g., "what is the current state of the area?").
- Density Estimation: This task requires a high-level spatial understanding of the urban layout. Models must classify the density of residential houses in the image as low, moderate, or high.
- Risk Assessment: Designed for urgent decision-support, these binary yes/no questions determine the immediate need for rescue intervention based on the severity of the visible flooding.

## IV. Methodology

Our approach follows a two-stage pipeline that combines in-context learning (ICL) with instruction-driven Chain-of-Thought (CoT) reasoning for post-disaster visual question answering. Given a target UAV image and question, exemplars are first retrieved and then integrated with a dual–multimodal large language model (MLLM) architecture, where one model generates task-specific reasoning instructions and a second model produces the final answer conditioned on the image. By varying how contextual exemplars are incorporated across the two stages, we evaluate different strategies for leveraging ICL to improve answer consistency and accuracy.

### A. Framework Architecture

For all ICL-based frameworks, visually relevant QA exemplars are first retrieved for each target image–question pair using a pretrained CLIP vision encoder to support effective in-context learning. All dataset embeddings are stored in a vector database for efficient retrieval. During inference, the embedding of the target image is compared against the stored image embeddings, and candidate exemplars are ranked according to their visual similarity. This similarity is computed using cosine similarity as follows:

$$\text{Sim}(I_{\text{target}}, I_i) = \frac{g(I_{\text{target}}) \cdot g(I_i)}{\|g(I_{\text{target}})\| \, \|g(I_i)\|}, \tag{1}$$

where $g(\cdot)$ denotes the embedding function, $I_{\text{target}}$ denotes the target image, and $I_i$ represents a candidate image from the retrieval set.

Based on this similarity ranking, exemplar selection is performed according to the question type. For non-counting (categorical) questions, exemplars are selected in a class-aware manner by retrieving one visually similar image per candidate answer option. This strategy ensures that the final ICL set covers all possible answers and reduces bias toward visually dominant classes. For counting-type questions, where answers are ordinal rather than categorical, the top two most visually similar images are selected directly from the ranked list and used as in-context examples.

Using these retrieved exemplars, we design and evaluate three distinct ICL-based frameworks to study how different ways of integrating ICL with CoT reasoning affect performance and hallucination behavior. All methods adopt a dual-MLLM pipeline inspired by prior work that separates reasoning from image-conditioned inference [10]. The Gemini-2.5-Flash model is used for all stages across all methods to ensure a fair comparison. All experiments are conducted in an inference-only setting, without any task-specific training or fine-tuning. As a result, performance differences across methods arise exclusively from differences in prompting strategy and ICL integration rather than from model optimization or hyperparameter tuning.

The first framework, Answer-ICL CoT (AIC), separates instruction generation and answer generation while applying ICL only at the second stage. In this setup, the first MLLM is provided solely with the question and a list of possible answers and is tasked with generating a step-by-step instruction or reasoning guideline. This instruction is then passed to a second MLLM together with the target image and the retrieved QA exemplars, which are used for in-context learning to produce the final answer.

Motivated by recent findings indicating that combining CoT and ICL within the same prompt can negatively impact accuracy [21], we propose Instruction-ICL CoT (IIC). In this framework, the retrieved QA exemplars are provided to the first MLLM alongside the question, enabling ICL during instruction generation. The resulting instruction is then passed to a second MLLM together with the target image, but without additional ICL examples. Here, ICL is used only in the first stage to guide the formation of more dataset-aware and task-consistent instructions.

The third framework, Both-ICL CoT (BIC), applies ICL at both stages of the pipeline. The same set of retrieved QA exemplars is provided to both the instruction-generation MLLM and the answer-generation MLLM. This allows in-context learning to influence both the reasoning structure and the final decision-making process.

TABLE I: VQA ACCURACY (%) COMPARISON ON THE FLOODNET DATASET ACROSS DIFFERENT METHODS.

| Question Type | IIC | AIC | BIC | zero-shot |
|---|---|---|---|---|
| Building Condition | **92.61** | 82.95 | 88.07 | 83.52 |
| Complex Counting | **28.80** | **28.80** | 27.75 | 24.08 |
| Density Estimation | 43.75 | 42.61 | **50.00** | 39.20 |
| Entire Condition | **95.64** | 88.14 | 90.56 | 90.80 |
| Risk Assessment | **94.32** | **94.32** | 92.05 | 91.48 |
| Road Condition | **92.77** | 89.55 | **92.77** | 89.15 |
| Simple Counting | **31.82** | 25.57 | **31.82** | 26.70 |

For all three methods, the set of possible answers is explicitly provided during instruction generation to constrain the reasoning space and help the second MLLM identify the correct output. Additionally, the instruction specifies the final answer to be enclosed within <*start*> and <*end*> tags, enabling reliable extraction and automated accuracy evaluation.

### *B. Results and Discussion*

Table I presents a quantitative comparison between the three proposed frameworks and a zero-shot baseline, allowing us to assess whether observed performance gains arise from the prompting strategy rather than solely from the strength of the underlying model. Overall, AIC consistently underperforms the other two methods across most question categories, and in some cases even falls below the zero-shot baseline.

We attribute this behavior primarily to the absence of ICL during instruction generation. Without exposure to exemplar QA pairs, the first-stage MLLM produces instructions that are less aligned with dataset-specific annotation patterns and labeling conventions. This limitation is illustrated in Fig.2, where the question *"What is the overall condition of the given image?"* is evaluated. In this example, although the image contains visible water, it is labeled as *non-flooded* in the dataset, reflecting how flooding is interpreted in relation to its impact on surrounding structures and infrastructure. For the AIC method, the generated instruction emphasizes generic criteria such as whether water extends beyond natural water bodies but fails to reference residential structures or roads. As a result, the instruction lacks key contextual cues required for correct classification. In contrast, both IIC and BIC, which incorporate ICL during instruction generation, explicitly learn from exemplars that flooding must involve encroachment upon residential areas. Their instructions include guidance such as searching for water submerging streets, driveways, or houses. During final answer generation, these methods correctly identify that although water is present in the image, it does not affect residential regions, leading to a correct non-flooded prediction.

Across all methods, binary questions such as yes/no and flooded/non-flooded (Table I, rows 1, 4, 5, and 6) achieve relatively high accuracy. In contrast, counting and density estimation questions perform worse due to limited numerical reasoning in VQA models [22], [23]. In FloodNet, density estimation depends on counting densely packed houses, which is challenging in aerial imagery. Additionally, we find that in several cases the model incorrectly reasons about vegetation density instead of residential density, leading to incorrect answers. In this regard, we believe that adding more exemplar QA pairs and incorporating query refinement could improve performance across all question types, which we leave for future studies. Notably, the proposed framework is general and can be applied to other disaster scenarios and VQA datasets with similar image–question structures. Finally, overall accuracy may be affected by annotation noise, which may slightly underestimate performance.

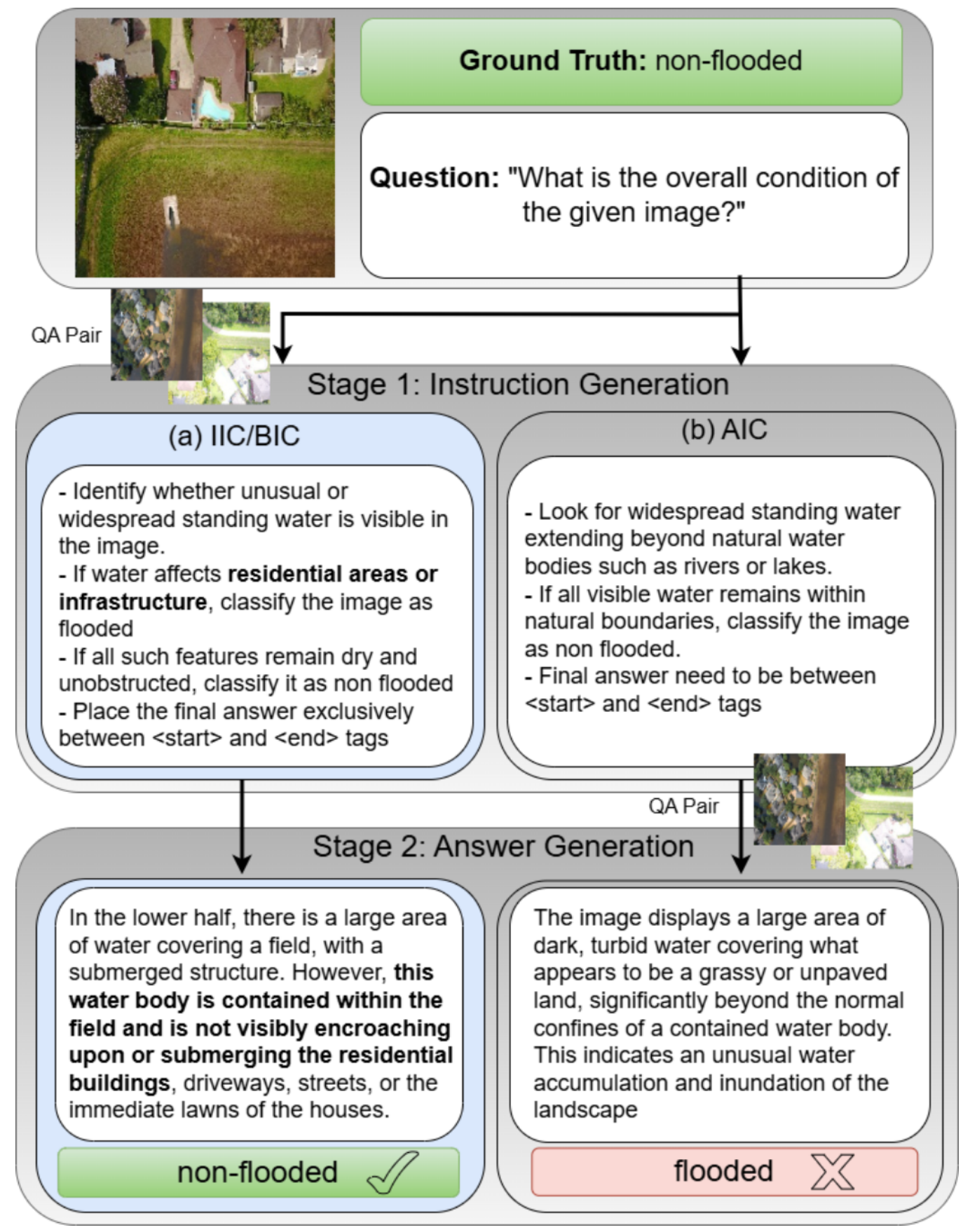

Fig. 2: Comparison of instruction generation and final rationale between AIC, IIC, and BIC methods

## V. CONCLUSION

This paper investigated how in-context learning can be integrated within instruction-driven Chain-of-Thought frameworks to improve post-disaster visual question answering using pre-trained multimodal large language models. By evaluating three dual-stage prompting strategies on the FloodNet dataset, we demonstrated that incorporating contextual exemplars during instruction generation leads to more consistent and accurate predictions. Our findings highlight the importance of structured reasoning and exemplar-aware instruction design for reliable UAV-based disaster assessment and motivate future work on more robust multimodal reasoning pipelines.

## ACKNOWLEDGMENT

This work was partially supported by the National Science Foundation under grants #2423211 and #2401942, the Consortium for Enhancing Resilience and Catastrophe Modeling.